\documentclass[letterpaper]{article} % DO NOT CHANGE THIS
\usepackage{aaai2026}  % DO NOT CHANGE THIS
\usepackage{times}  % DO NOT CHANGE THIS
\usepackage{helvet}  % DO NOT CHANGE THIS
\usepackage{courier}  % DO NOT CHANGE THIS
\usepackage[hyphens]{url}  % DO NOT CHANGE THIS
\usepackage{graphicx} % DO NOT CHANGE THIS
\usepackage{natbib}  % DO NOT CHANGE THIS AND DO NOT ADD ANY OPTIONS TO IT
\usepackage{caption} % DO NOT CHANGE THIS AND DO NOT ADD ANY OPTIONS TO IT
\usepackage{algorithm}
\usepackage{algpseudocode}
 \usepackage{textcomp}
\usepackage{algorithm}
\usepackage{graphicx}
\usepackage{amsfonts}
\usepackage{amsmath}
\usepackage{threeparttable}
\usepackage{enumitem}
\usepackage{booktabs}
\usepackage{algpseudocode}
\usepackage{newfloat}
\usepackage{listings}
\usepackage{multirow}
\usepackage{times}
\usepackage{helvet}
\usepackage{courier}
\usepackage{xcolor}
\makeatletter\def\@listi{\leftmargin\leftmargini \topsep .5em \parsep .5em \itemsep .5em}
\def\@listii{\leftmargin\leftmarginii \labelwidth\leftmarginii \advance\labelwidth-\labelsep \topsep .4em \parsep .4em \itemsep .4em}
\def\@listiii{\leftmargin\leftmarginiii \labelwidth\leftmarginiii \advance\labelwidth-\labelsep \topsep .4em \parsep .4em \itemsep .4em}\makeatother

\newcounter{checksubsection}
\newcounter{checkitem}[checksubsection]

\usepackage{newfloat}
\usepackage{listings}
\DeclareCaptionStyle{ruled}{labelfont=normalfont,labelsep=colon,strut=off} % DO NOT CHANGE THIS
\floatstyle{ruled}
\newfloat{listing}{tb}{lst}{}
\floatname{listing}{Listing}
\title{Spatial-Temporal Feedback Diffusion Guidance for Controlled Traffic Imputation}
\author{
    Xiaowei Mao\textsuperscript{\rm 1}\textsuperscript{*},
    ~Huihu Ding\textsuperscript{\rm 1}\textsuperscript{*},
    ~Yan Lin\textsuperscript{\rm 4},
    ~Tingrui Wu\textsuperscript{\rm 1},\\
    ~Shengnan Guo\textsuperscript{\rm 1, 2},
    ~Dazhuo Qiu\textsuperscript{\rm 4},
    ~Feiling Fang\textsuperscript{\rm 5},
    ~Jilin Hu\textsuperscript{\rm 6},
    ~Huaiyu Wan\textsuperscript{\rm 1, 3}\textsuperscript{\(\dagger\)} % 只保留名字和手动符号，不要加 thanks
}
 \affiliations {
     \textsuperscript{\rm 1}School of Computer Science and Technology, Beijing Jiaotong University, China\\
     \textsuperscript{\rm 2}Key Laboratory of Big Data \& Artificial Intelligence in Transportation, Ministry of Education, China\\
    \textsuperscript{\rm 3}Beijing Key Laboratory of Traffic Data Mining and Embodied Intelligence, China \\
    \textsuperscript{\rm 4}Department of Computer Science, Aalborg University, Denmark\\
     \textsuperscript{\rm 5}School of Artificial Intelligence, China University of Geoscience, Beijing, China\\
     \textsuperscript{\rm 6}School of Data Science and Engineering, East China Normal University, China\\
    \{maoxiaowei, 22231044, guoshn, hywan\}@bjtu.edu.cn, lyan@cs.aau.dk, jlhu@dase.ecnu.edu.cn\\
}
  
\begin{document}

\maketitle
\begingroup
 
    \renewcommand{\thefootnote}{}
    
    \makeatletter
    \renewcommand\@makefntext[1]{\noindent #1} 
    \makeatother
    \footnotetext{\textsuperscript{\(\dagger\)}Corresponding author.}
    \footnotetext{\textsuperscript{*}These authors contributed equally.}
\endgroup

\begin{abstract}
Imputing missing values in spatial-temporal traffic data is essential for intelligent transportation systems. Among advanced imputation methods, score-based diffusion models have demonstrated competitive performance. These models generate data by reversing a noising process, using observed values as conditional guidance. However, existing diffusion models typically apply a uniform guidance scale across both spatial and temporal dimensions, which is inadequate for nodes with high missing data rates. Sparse observations provide insufficient conditional guidance, causing the generative process to drift toward the learned prior distribution rather than closely following the conditional observations, resulting in suboptimal imputation performance.

To address this, we propose FENCE, a spatial-temporal feedback diffusion guidance method designed to adaptively control guidance scales during imputation. First, FENCE introduces a dynamic feedback mechanism that adjusts the guidance scale based on the posterior likelihood approximations. The guidance scale is increased when generated values diverge from observations and reduced when alignment improves, preventing overcorrection. Second, because alignment to observations varies across nodes and denoising steps, a global guidance scale for all nodes is suboptimal. FENCE computes guidance scales at the cluster level by grouping nodes based on their attention scores, leveraging spatial-temporal correlations to provide more accurate guidance. Experimental results on real-world traffic datasets show that FENCE significantly enhances imputation accuracy.

\end{abstract}

\begin{links}
    \link{Code}{https://github.com/maoxiaowei97/FENCE}

\end{links}

\section{Introduction}
Spatial-temporal traffic data are often represented as a graph of spatial-temporal time series, where each node is a traffic sensor continuously collecting observations and edges describe sensor relationships~\cite{guo2024experimental}. Traffic data is essential for Intelligent Transportation Systems (ITS), supporting key services in ITS, such as real-time traffic display, traffic prediction, and traffic signal control. However, this data is frequently incomplete due to equipment malfunctions, and network failures. These missing values degrade the performance of dependent applications. Consequently, imputing missing values is essential to ensure data quality and reliability~\cite{wang2024deep, miao2022experimental}.
\begin{figure}[!t]

    \centering
    \includegraphics[width= 1\linewidth]{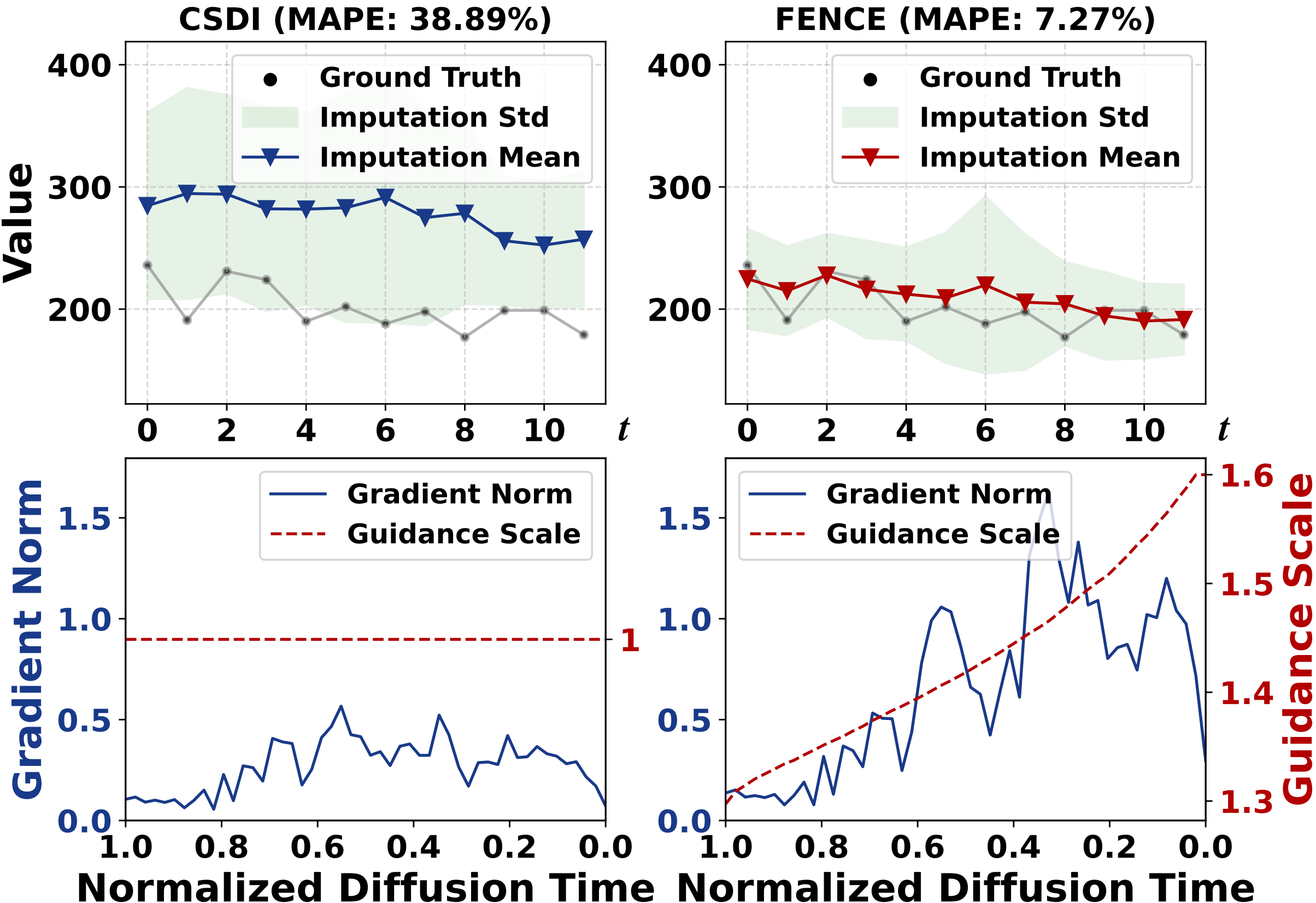}

        \caption{Motivation for FENCE. Unlike CSDI, which uses a fixed guidance scale, FENCE dynamically adjusts guidance scale based on consistency with observed data. }
        
    \label{fig:intro}

\end{figure}
 
\par Deep learning paradigms for spatial-temporal imputation can be broadly categorized into two main approaches: discriminative and generative models. Discriminative models directly learn a mapping function from observed to missing values using architectures like Recurrent Neural Networks (RNNs)~\cite{miao2021generative}, Graph Neural Networks (GNNs)~\cite{lao2022variational}, and Transformers~\cite{nie2024imputeformer}. While often straightforward to train, their focus on direct prediction limits their ability to capture data distributions and the uncertainty associated with missing values.

In contrast, generative models aim to learn the underlying probability distribution of the data, enabling high-fidelity imputations~\cite{ijcai2019p429, yoon2018gain}. Imputation is formulated as a conditional generation task, where missing values are sampled from this learned distribution based on the observed data~\cite{zhou2024mtsci}. Among these, score-based diffusion models~\cite{song2020score} have emerged as a competitive method for imputation. These models learn the score function, defined as the gradient of the data's log-likelihood, and utilize the conditional score to guide the generation process.

However, these models often yield suboptimal performance, particularly for nodes with high missing data rates. As illustrated in Fig.~\ref{fig:intro}, a node with no observations during a time period is imputed inaccurately by CSDI~\cite{tashiro2021csdi}, where even the estimated lower and upper bounds may fail to encompass the ground truth. This issue can be quantitatively assessed by examining the generative process. The degree to which the generated data \(\boldsymbol{x}_k\) satisfies condition \(\boldsymbol{c}\) is measured by the posterior likelihood, \(p_{\theta, k}(\boldsymbol{c}|\boldsymbol{x}_k)\). To improve this likelihood, the diffusion model is guided by the gradient of the log-posterior likelihood. This guidance term is approximated by the difference between the conditional and unconditional score functions: \(\nabla_{\boldsymbol{x}_k}{\log}p_{\theta, k}(\boldsymbol{x}_k|\boldsymbol{c}) - \nabla_{\boldsymbol{x}_k}{\log}p_{\theta, k}(\boldsymbol{x}_k)\). The L2-norm of this gradient vector quantifies the guidance strength. As shown in Fig.~\ref{fig:intro}, during the generation process of CSDI, the node without observations exhibits a consistently low gradient norm. This indicates that the learned conditional distribution has collapsed to the unconditional prior. Consequently, the generative process is biased towards sampling from the marginal distribution \(p_{\theta, k}(\boldsymbol{x}_k)\) instead of the conditional distribution \(p_{\theta, k}(\boldsymbol{x}_k|\boldsymbol{c})\). Existing diffusion models for imputation lack mechanisms to control the guidance strength, leading to insufficient adherence to specific conditional observations and suboptimal performance.

To address this issue, we propose FENCE (Spatial-Temporal \underline{FE}edback Diffusion Guida\underline{NCE}), a novel method for controlled traffic imputation that dynamically adjusts the guidance scales throughout the generative process. FENCE introduces a feedback mechanism that adjusts the guidance scale based on an approximation of the posterior likelihood. Specifically, when the posterior likelihood decreases, indicating that the generated values do not sufficiently adhere to the conditional observations, the guidance scale is increased to enhance alignment with the observations. In contrast, when the posterior likelihood is high, indicating good alignment between the generated values and the observations, the guidance scale is reduced to avoid overcorrection. Furthermore, to account for varying degrees of alignment with conditional observations across different nodes, FENCE computes the guidance scale at the cluster level. By leveraging spatial-temporal correlations, FENCE ensures more accurate guidance scale adjustments for nodes with limited data availability, thereby improving the imputation quality.

Our contributions are summarized as follows:

\begin{itemize}
    \item We propose FENCE, a spatial-temporal feedback diffusion guidance method that dynamically controls the guidance scales during the generative process, ensuring high-fidelity imputation of missing traffic data.
    \item We propose a cluster-aware guidance mechanism that leverages spatial-temporal correlations to compute accurate guidance scales tailored to each node.
    \item  Extensive experiments show that FENCE significantly enhances the imputation accuracy in real-world spatial-temporal traffic datasets.
    
\end{itemize}

\section{Related Work}

\par \noindent \textbf{Spatial-Temporal Imputation}. 
Spatial-temporal imputation methods can be broadly classified into discriminative and generative paradigms. Discriminative models~\cite{cao2018brits, che2018recurrent, ijcai2025p386}, such as SAITS~\cite{du2023saits} and ImputeFormer~\cite{nie2024imputeformer}, learn deterministic mappings from observed data but fail to explicitly model data distributions.

In contrast, generative models aim to learn the underlying data distribution and treat imputation as conditional sampling, generating plausible values for the missing entries given the observed data~\cite{yoon2018gain, fortuin2020gp, ipsen2022deal}. 

Score-based diffusion models are powerful generative models for imputation. These models learn the score function, which is the gradient of the log-likelihood of the data distribution. During imputation, they leverage the score of the conditional distribution to estimate missing values. Models such as CSDI \cite{tashiro2021csdi} and MIDM \cite{wang2023observed} condition the diffusion process on available observations. Several extensions further enhance conditioning: LSCD~\cite{fons2025lscd} incorporates spectral information; and PriSTI~\cite{liu2023pristi} integrates geographic context. To improve imputation consistency and inference speed, CSBI~\cite{chen2023provably} leverages a Schrödinger bridge formulation; MTSCI~\cite{zhou2024mtsci} generates multiple masks and auxiliary conditions during training; DSDI~\cite{xiao2025boundary} incorporates the predicted values into the denoising process, and CoSTI~\cite{solis2025costi} employs consistency training to reduce inference times.

Despite their effectiveness in modeling complex distributions, diffusion models face challenges in spatial-temporal imputation, especially for nodes with high missing rates. In such cases, the limited observed data may be insufficient to effectively guide the model from its learned prior to converge to the true conditional distribution. Consequently, the imputed values often reflect general data patterns rather than adhering to the available observations. A key limitation is that current models lack mechanisms to dynamically adjust the scales of guidance strength based on the observed data.

\section{Preliminaries}
% In this section, we provide the relevant definitions and formalize the problem of spatial-temporal traffic imputation.

\par \noindent {\textbf{Definition\hspace{3px}1.}} \textit{\textbf{Traffic Network.}} 
We define the traffic network as a graph, i.e., $G = (V, E, \mathbf{A})$, where $V$ represents the set of $|V|=N$ nodes (e.g., traffic sensors). $E$ represents the set of edges. $\mathbf{A} \in \mathbb{R}^{N \times N}$ is the adjacency matrix.

\par \noindent {\textbf{Definition\hspace{3px}2.}} \textit{\textbf{Traffic Data.}} Let \( x_{v,t} \in \mathbb{R} \) denote the traffic data observed at node \( v \in V \) at time slice \( t \). The traffic data at time \( t \) across all nodes is \( \boldsymbol{x}_t = (x_{1,t}, x_{2,t}, \dots, x_{N,t}) \in \mathbb{R}^N \), and the traffic data over \( T \) time slices is \( \boldsymbol{x} = (\boldsymbol{x}_1, \boldsymbol{x}_2, \dots, \boldsymbol{x}_T) \in \mathbb{R}^{N \times T} \).

\par \noindent {\textbf{Definition\hspace{3px}3.}} \textit{\textbf{Mask Matrix.}}
To indicate the missing position in the observed traffic data, we introduce an observation masking matrix $\mathbf{M} \in \mathbb{R}^{N \times T}$, where $m_{v,t}=0$ when $x_{v,t}$ is missing, and $m_{v,t}=1$ when $x_{v,t}$ is observed. The observed values in \(\boldsymbol{x}\) are denoted as \(\boldsymbol{x}^{o} = \boldsymbol{x} \odot \mathbf{M}\), and the missing values in \(\boldsymbol{x}\) are denoted as \(\boldsymbol{x}^{m} = \boldsymbol{x} \odot (\mathbf{1} - \mathbf{M})\).

\par \noindent {\textbf{Spatial-Temporal Traffic Imputation.}} Given the incomplete traffic observations $\boldsymbol{x}$, the mask matrix $\mathbf{M} \in \mathbb{R}^{N \times T}$ over \(T\) time slices, and a network graph $G$, the objective is to estimate the missing values in \(\boldsymbol{x}\) such that the estimation error at the missing positions is minimized.

\section{Methods}

This section presents our controlled traffic imputation method, beginning with guidance in diffusion models for imputation. We then propose spatial-temporal feedback diffusion guidance, followed by the theoretical foundations, key mechanisms, and procedures for training and inference.

\begin{figure*}[hbtp]
		\centering
		\includegraphics[width=1\linewidth]{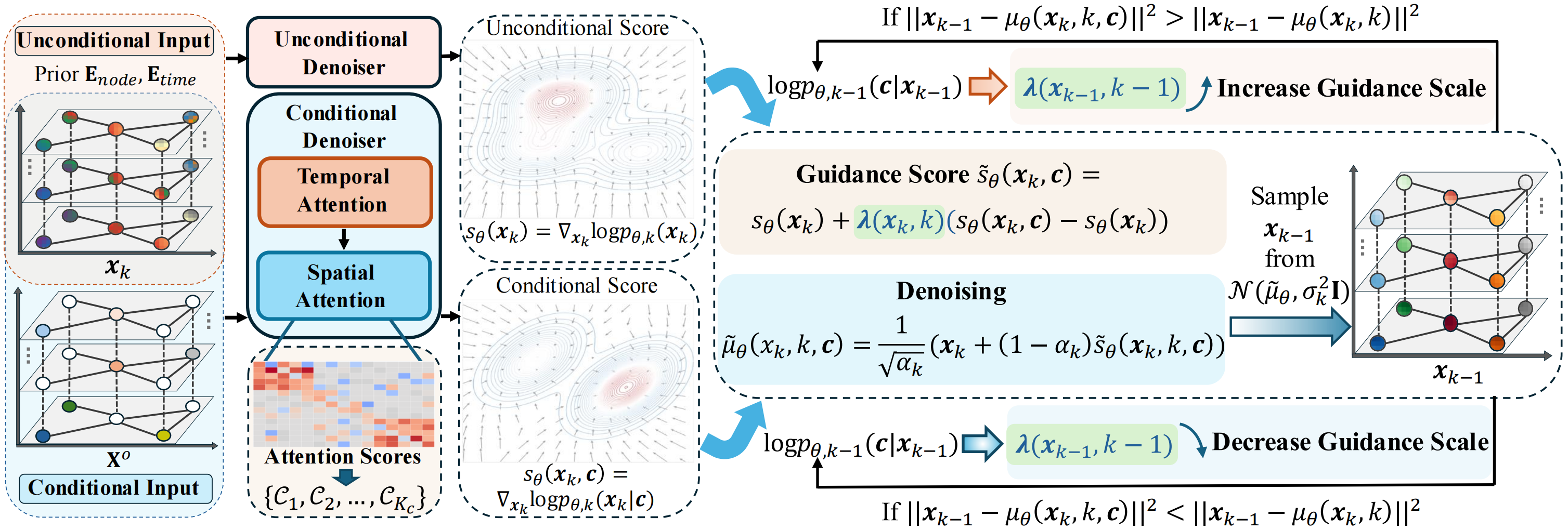}

   \caption{
   FENCE performs imputation by estimating both conditional and unconditional scores. It dynamically adjusts the guidance scale at each step by evaluating the posterior likelihood, controlling the scales of the conditional guidance strength to ensure consistency with observed data.}

		\label{fig_model}
\end{figure*}

\subsection{Guidance in Diffusion Models for Traffic Imputation}

In spatial-temporal diffusion models for traffic imputation, guidance is achieved by conditioning the reverse process on observed data and structural priors. This conditional information $\boldsymbol{c}$, is produced by a conditioning network $\mathcal{F}_{cond}$, which captures temporal and spatial dependencies. Given the observed data $\boldsymbol{x}^o$, the network first employs temporal attention to capture dependencies across time for each node, followed by spatial attention to aggregate information across nodes for each time slice. The conditioning also incorporates structural priors, such as node embeddings $\mathbf{E_{\text{node}}} \in \mathbb{R}^{N \times d}$ and learnable time slice embeddings $\mathbf{E_{\text{time}}} \in \mathbb{R}^{T \times d}$. The resulting vector, \(\boldsymbol{c} = \mathcal{F}_{cond}(\boldsymbol{x}^o, \mathbf{E_{\text{node}}}, \mathbf{E_{\text{time}}})\), then guides the reverse process. Starting from Gaussian noise $\boldsymbol{x}_K \sim \mathcal{N}(\mathbf{0}, \mathbf{I})$, the model iteratively denoises the sample, with each step of the reverse process conditioned on $\boldsymbol{c}$:
\begin{equation}
p_{\theta}(\boldsymbol{x}_{k-1} | \boldsymbol{x}_k, \boldsymbol{c}) = \mathcal{N}(\boldsymbol{x}_{k-1}; {\mu}_{\theta}(\boldsymbol{x}_k, k, \boldsymbol{c}), \sigma_k^2 \mathbf{I}),
\end{equation}
where the mean ${\mu}_{\theta}$ is parameterized using a denoising network ${\epsilon}_{\theta}$ that predicts the added noise at step $k$:

\begin{equation}
{\mu}_{\theta}(\boldsymbol{x}_k, k, \boldsymbol{c}) = \frac{1}{\sqrt{\alpha_k}} \left( \boldsymbol{x}_k - \frac{1-\alpha_k}{\sqrt{1-\bar{\alpha}_k}} {\epsilon}_{\theta}(\boldsymbol{x}_k, k, \boldsymbol{c}) \right)
\end{equation}
The denoising network \({\epsilon}_{\theta}\) is trained to learn the conditional score function, where the score is proportional to the predicted noise: \(\nabla_{\boldsymbol{x}_k} \log p_{\theta, k}(\boldsymbol{x}_k | \boldsymbol{c}) = -\frac{1}{\sqrt{1-\bar{\alpha}_k}} {\epsilon}_{\theta}(\boldsymbol{x}_k, k, \boldsymbol{c})\). This score function provides the gradient direction to iteratively guide the sample \(\boldsymbol{x}_k\) to maximize its likelihood under the learned conditional distribution.

While the objective of a diffusion model is to maximize the data likelihood, this does not ensure the maximization of the posterior likelihood, $p_{\theta, k}(\boldsymbol{c}|\boldsymbol{x}_k)$, which quantifies how well a sample satisfies the observations. To better align the generated data with the observations, the reverse process can be guided by the gradient of the log-posterior likelihood, $\nabla_{\boldsymbol{x}_k} \log p_{\theta, k}(\boldsymbol{c}|\boldsymbol{x}_k)$, which can be formulated in terms of learnable components using Bayes' theorem:

\begin{equation}
\nabla_{\boldsymbol{x}_k} \log p_{\theta, k}(\boldsymbol{c}|\boldsymbol{x}_k) = \nabla_{\boldsymbol{x}_k}{\log}p_{\theta, k}(\boldsymbol{x}_k|\boldsymbol{c}) - \nabla_{\boldsymbol{x}_k}{\log}p_{\theta, k}(\boldsymbol{x}_k)
\label{eq:bayes_score}
\end{equation}
This equation reveals that the guidance can be achieved by subtracting the unconditional score, $\nabla_{\boldsymbol{x}_k}{\log}p_{\theta, k}(\boldsymbol{x}_k)$, from the conditional score. Classifier Free Guidance (CFG)~\cite{ho2022classifier} provides an efficient implementation by training a single denoising network to learn both scores. This is achieved by randomly providing the network during training with either the conditioning vector or an unconditional vector. The latter is constructed to model the prior distribution by feeding the conditioning network empty observations while retaining the structural priors. Using these scores, CFG constructs a guidance score by leveraging the gradient of the log-posterior likelihood. Specifically, the guidance score, $\nabla \log \tilde{p}_{\theta, k}(\boldsymbol{x}_k|\boldsymbol{c})$, is constructed by scaling this gradient (as derived in Eq.~\ref{eq:bayes_score}) by a guidance scale $\lambda$ and adding it to the unconditional score:
\begin{equation}
\begin{split}
    \nabla_{\boldsymbol{x}_k} \log \tilde{p}_{\theta, k}(\boldsymbol{x}_k|\boldsymbol{c}) ={}& \nabla_{\boldsymbol{x}_k} \log p_{\theta, k}(\boldsymbol{x}_k) \\
    & + \lambda \big( \nabla_{\boldsymbol{x}_k} \log p_{\theta, k}(\boldsymbol{x}_k|\boldsymbol{c}) \\
    & - \nabla_{\boldsymbol{x}_k} \log p_{\theta, k}(\boldsymbol{x}_k) \big)
\end{split}
\end{equation}

The guidance scale $\lambda$ is a hyperparameter that adjusts the strength of the conditioning signal.

\subsection{Spatial-Temporal Feedback Diffusion Guidance}
\label{ST-FBG}

CFG applies a uniform guidance scale $\lambda$ to all nodes across all denoising steps. This approach has two limitations for traffic imputation. First, $\lambda$ is a fixed hyperparameter that is difficult to optimize. Second, it fails to account for the varying degrees to which imputed values for different nodes satisfy the conditional observations at different times.

To address this, we introduce a feedback guidance mechanism that adaptively adjusts the guidance scales based on the posterior likelihood at each denoising step \(k\). The posterior likelihood quantifies the alignment between the generated sample \(\boldsymbol{x}_k\) and the condition \(\boldsymbol{c}\). When the posterior likelihood is high, indicating good alignment with the observed data, then the guidance scale remains low to avoid overcorrection. In contrast, when the posterior likelihood decreases, the guidance scale is increased to ensure adherence to the conditional information.

\subsubsection{Posterior-Driven Dynamic Guidance Scaling}
To enable controlled imputation, we begin by introducing a global guidance mechanism. This approach treats the traffic data matrix comprising all nodes over $T$ time slices as a sample. At each denoising step $k$, a unified guidance scale, denoted as $\lambda(\boldsymbol{x}_k, k)$, is dynamically adjusted for the corresponding noised sample $\boldsymbol{x}_k$ based on the posterior likelihood. This scale controls the guidance vector, which is the difference between the conditional and unconditional score estimates. We define the unconditional score as $s_{\theta}(\boldsymbol{x}_k) := \nabla_{\boldsymbol{x}_k} \log p_{\theta, k}(\boldsymbol{x}_k)$ and the conditional score as $s_{\theta}(\boldsymbol{x}_k, \boldsymbol{c}) := \nabla_{\boldsymbol{x}_k} \log p_{\theta, k}(\boldsymbol{x}_k|\boldsymbol{c})$. The resulting guidance score is:
\begin{equation}
\begin{split}
    \nabla_{\boldsymbol{x}_k} \log \tilde{p}_{\theta, k}(\boldsymbol{x}_{k}|\boldsymbol{c}) ={}& s_{\theta}(\boldsymbol{x}_{k}) \\
    & + \lambda(\boldsymbol{x}_{k}, k) \left( s_{\theta}(\boldsymbol{x}_k, \boldsymbol{c}) - s_{\theta}(\boldsymbol{x}_{k}) \right)
\end{split}
\label{eq:fbg_form}
\end{equation}

To compute the guidance scale, we adopt the additive error formulation~\cite{koulischer2025feedback}. This formulation assumes that the learned conditional distribution $p_{\theta,k}(\boldsymbol{x}_{k}|\boldsymbol{c})$ is a linear combination of the true conditional distribution $p_k(\boldsymbol{x}_{k}|\boldsymbol{c})$ and the unconditional distribution $p_k(\boldsymbol{x}_{k})$:
\begin{equation}
p_{\theta,k}(\boldsymbol{x}_{k}|\boldsymbol{c}) = (1 - \pi)p_{k}(\boldsymbol{x}_{k}) + \pi p_{k}(\boldsymbol{x}_{k}|\boldsymbol{c}),
\label{eq:additive_error}
\end{equation}
where $\pi \in [0, 1]$ is a hyperparameter that indicates the prior confidence in how well the conditional generative model has learned to adhere to the condition \(\boldsymbol{c}\). Next, we compute the score function of this target distribution. Applying the chain rule for logarithmic derivatives yields (see Appendix for the full derivation):
\begin{equation}
    \nabla_{\boldsymbol{x}_k} \log p_{\theta,k}(\boldsymbol{x}_k|\boldsymbol{c}) = \frac{\nabla_{\boldsymbol{x}_k} \left( p_{\theta,k}(\boldsymbol{x}_k|\boldsymbol{c}) - (1 - \pi)p_{\theta,k}(\boldsymbol{x}_k) \right)}{p_{\theta,k}(\boldsymbol{x}_k|\boldsymbol{c}) - (1 - \pi)p_{\theta,k}(\boldsymbol{x}_k)}
\end{equation}
By formulating the probability densities in the numerator and denominator using Bayes' theorem and applying the score identity ($\nabla_{\boldsymbol{x}_k} p(\boldsymbol{x}_k) = p(\boldsymbol{x}_k)\nabla_{\boldsymbol{x}_k} \log p(\boldsymbol{x}_k)$), this expression can be arranged into the guidance score formulation of Eq.~\ref{eq:fbg_form} (see Appendix). This process yields the formulation for the guidance scale:
\begin{equation}
\lambda(\boldsymbol{x}_k, k) = \frac{p_{\theta,k}(\boldsymbol{c}|\boldsymbol{x}_k) / p_{\theta,k}(\boldsymbol{c})}{p_{\theta,k}(\boldsymbol{c}|\boldsymbol{x}_k) / p_{\theta,k}(\boldsymbol{c}) - (1 - \pi)}
\label{eq:lambda_full}
\end{equation}
In practice, assuming the prior $p_{\theta,k}(\boldsymbol{c})$ is constant yields:
\begin{equation}
    \lambda(\boldsymbol{x}_k, k) \approx \frac{p_{\theta,k}(\boldsymbol{c}|\boldsymbol{x}_k)}{p_{\theta,k}(\boldsymbol{c}|\boldsymbol{x}_k) - (1-\pi)}
    \label{eq:lambda_practical}
\end{equation}
This formulation indicates that the guidance scale is a function of the posterior likelihood $p_{\theta,k}(\boldsymbol{c}|\boldsymbol{x}_k)$, and achieves our objective: when the posterior is high, indicating high consistency with the condition \(\boldsymbol{c}\), the guidance scale approaches 1. As the posterior likelihood decreases toward the threshold $(1-\pi)$, the guidance scale increases, applying stronger guidance to ensure adherence to the conditional distribution.

\subsubsection{Posterior Likelihood Estimation}
The formulation for $\lambda(\boldsymbol{x}_k, k)$ in Eq.~\ref{eq:lambda_practical} depends on the posterior likelihood, which is not directly accessible. Inspired by~\cite{koulischer2025feedback}, we can estimate this value by tracking the diffusion's reverse Markov chain. The derivation begins with the definition of the posterior, $p_{\theta, k-1}(\boldsymbol{c}|\boldsymbol{x}_{k-1:K})$. By applying the chain rule of probability and leveraging the Markov property of the reverse diffusion process (i.e., $p_{\theta}(\boldsymbol{x}_{k-1}|\boldsymbol{x}_{k:K}, \boldsymbol{c}) = p_{\theta}(\boldsymbol{x}_{k-1}|\boldsymbol{x}_{k}, \boldsymbol{c})$), the posterior at step $k-1$ is:

\begin{equation}
p_{\theta, k-1}(\boldsymbol{c}|\boldsymbol{x}_{k-1}) = p_{\theta, k}(\boldsymbol{c}|\boldsymbol{x}_k) \cdot \frac{p_{\theta}(\boldsymbol{x}_{k-1}|\boldsymbol{x}_k, \boldsymbol{c})}{p_{\theta}(\boldsymbol{x}_{k-1}|\boldsymbol{x}_k)}
\label{eq: posterior_multiple_relation}
\end{equation}
Taking the logarithm of Eq.~\ref{eq: posterior_multiple_relation}, we can obtain the update function of the posterior likelihood from step \(k\) to \(k-1\):
\begin{equation}
\begin{split}
\log p_{\theta, k-1}(\boldsymbol{c}|\boldsymbol{x}_{k-1}) ={}& \log p_{\theta, k}(\boldsymbol{c}|\boldsymbol{x}_k) + \log p_{\theta}(\boldsymbol{x}_{k-1}|\boldsymbol{x}_k, \boldsymbol{c}) \\
& - \log p_{\theta}(\boldsymbol{x}_{k-1}|\boldsymbol{x}_k)
\end{split}
\label{posterior_update}
\end{equation}
As the reverse transition distributions are Gaussian, the log-likelihood difference becomes:
\begin{equation}
\begin{split}
& \log p_{\theta}(\boldsymbol{x}_{k-1}|\boldsymbol{x}_k, \boldsymbol{c}) - \log p_{\theta}(\boldsymbol{x}_{k-1}|\boldsymbol{x}_k) \\
& \qquad = \frac{1}{2\sigma_k^2} \left( \|\boldsymbol{x}_{k-1} - {\mu}_{\theta}(\boldsymbol{x}_k)\|^2 - \|\boldsymbol{x}_{k-1} - {\mu}_{\theta}(\boldsymbol{x}_k|\boldsymbol{c})\|^2 \right)
\end{split}
\end{equation}
This formulation enables updating the posterior likelihood by comparing the outputs of the conditional and unconditional models at each step. Additionally, we introduce two hyperparameters: a temperature $\tau$ to scale the update strength and an offset $\delta$ to ensure guidance activates properly in early diffusion stages. This leads to the parameterized update equation for posterior:
\begin{multline} \label{eq:posterior_update_practical}
    \log p_{\theta, k-1}(\boldsymbol{c}|\boldsymbol{x}_{k-1}) = \log p_{\theta, k}(\boldsymbol{c}|\boldsymbol{x}_k) \\
    - \frac{\tau}{2\sigma_k^2} \Bigl( \|\boldsymbol{x}_{k-1} - {\mu}_{\theta}(\boldsymbol{x}_k|\boldsymbol{c})\|^2 - \|\boldsymbol{x}_{k-1} - {\mu}_{\theta}(\boldsymbol{x}_k)\|^2 \Bigr) - \delta
\end{multline}
By initializing $\log p_{\theta, K}(\boldsymbol{c}|\boldsymbol{x}_T)$ (e.g., to 0, assuming a uniform prior distribution) and applying this update rule iteratively from $k=K$ to $1$, we can estimate the log-posterior at each denoising step. The feedback guidance loop is thus complete: at each step $k$, we use the current guidance scale $\lambda(\boldsymbol{x}_k, k)$ to sample $\boldsymbol{x}_{k-1}$. We then use this new sample $\boldsymbol{x}_{k-1}$ in Eq.~\ref{eq:posterior_update_practical} to update the posterior $p_{\theta, k-1}(\boldsymbol{c}|\boldsymbol{x}_{k-1})$. Finally, this new posterior is fed into Eq.~\ref{eq:lambda_practical} to determine the guidance scale $\lambda(\boldsymbol{x}_{k-1}, k-1)$ for the next step.

\subsubsection{Cluster-Aware Feedback Guidance}
\label{sec:cluster_guidance}
While the global guidance mechanism adapts the scale across denoising steps, it applies this scale uniformly to all nodes, which is suboptimal because nodes differ in their alignment to observations. Fully per-node scaling, however, can be statistically unstable under sparse observations. To address this, we introduce a cluster-aware feedback guidance strategy, which aggregates information from a group of correlated nodes to compute the guidance scale for each node. To group the nodes, we leverage the spatial attention scores, $\mathbf{A}_{attn} \in \mathbb{R}^{N \times N}$, from the conditional denoising network. Since the attention scores quantify dynamic correlations that evolve during the reverse process, we employ k-means clustering at each denoising step to partition the set of nodes $V$ into $K_c$ disjoint clusters, $\{\mathcal{C}_1, \mathcal{C}_2, \dots, \mathcal{C}_{K_c}\}$.

During each step of the reverse diffusion process, for any node $i$ belonging to the current cluster $\mathcal{C}_j$, we compute a cluster-level log-posterior. The aggregation rule for the cluster-level log-posterior is defined as:
\begin{equation}
\log p_{\theta, k-1, \mathcal{C}_j}(\boldsymbol{c}|\boldsymbol{x}_{k-1}) = \frac{1}{|\mathcal{C}_j|} \sum_{l \in \mathcal{C}_j} \log p_{\theta, k-1, l}(\boldsymbol{c}|\boldsymbol{x}_{k-1}),
\label{eq:cluster_posterior}
\end{equation}
where $\log p_{\theta, k-1, l}(\boldsymbol{c}|\boldsymbol{x}_{k-1})$ is the individually updated log-posterior for node $l$ using Eq.~\ref{eq:posterior_update_practical}. By averaging over all nodes in the cluster, we obtain a more stable estimate that is less susceptible to the high variance from any single node.

The cluster-level posterior, $p_{\theta, k, \mathcal{C}_j}(\boldsymbol{c}|\boldsymbol{x}_k)$, is then used to compute a shared guidance scale for all nodes within that cluster, using the formulation from Eq.~\ref{eq:lambda_practical}:
\begin{equation}
    \lambda_{\mathcal{C}_j}(\boldsymbol{x}_k, k) = \frac{p_{\theta, k, \mathcal{C}_j}(\boldsymbol{c}|\boldsymbol{x}_k)}{p_{\theta, k, \mathcal{C}_j}(\boldsymbol{c}|\boldsymbol{x}_k) - (1-\pi)}
    \label{eq:cluster_scale}
\end{equation}

\subsection{Training and Inference}\label{Train-Inference}

\subsubsection{Training.}
FENCE requires both unconditional and conditional predictions to compute the guidance scales. To prevent the learning of the unconditional prior from interfering with the conditional imputation, we adopt a two-stage training procedure. First, we train an unconditional generative model to learn the prior distribution $p_{\theta}(\boldsymbol{x})$. In this stage, the denoising network ${\epsilon}_{\theta}$ is trained using only the unconditional vector which is generated from structural priors without any observations. After convergence, the weights of this unconditional model are saved. Next, we fine-tune this pre-trained model for the conditional imputation. The network weights are initialized from the saved unconditional model. The model is then trained using the conditional observations.

\subsubsection{Inference.}
During inference, the denoising network utilizes a conditional context $\boldsymbol{c}$ and an unconditional context $\boldsymbol{c}_{\text{uncond}}$ as inputs. Instead of applying a fixed guidance scale, FENCE dynamically adjusts the guidance scale at each step of the denoising process. This adjustment is driven by a feedback loop that continuously estimates the posterior likelihood to assess the alignment between the current sample and the conditional observations. Furthermore, to account for the varying degrees of alignment with conditional observations across different nodes, the feedback is computed at a cluster level, leveraging spatial-temporal correlations. The inference procedure is presented in Algorithm~\ref{alg:fence_inference}.

\begin{algorithm}[t]
\caption{Inference of FENCE}
\label{alg:fence_inference}
\begin{algorithmic}[1]
\State \textbf{Input:} Conditional network ${\epsilon}_{\theta}$, unconditional network ${\epsilon}_{\theta}^{\text{uncond}}$, observed data $\boldsymbol{x}^o$ and mask $\mathbf{M}$, total denoising steps $K$, hyperparameters $\pi, \tau, \delta$, $K_c$.

\State \textbf{Initialize:}
\State Sample $\boldsymbol{x}_K \sim \mathcal{N}(0, \mathbf{I})$.
\State Initialize $\log p_{\theta, K, i}(\boldsymbol{c}|\boldsymbol{x}_K) \leftarrow 0$ for all nodes.

\For{$k = K, \dots, 1$}
    \State ${\epsilon}_{\text{cond}} \leftarrow {\epsilon}_{\theta}(\boldsymbol{x}_k, k, \boldsymbol{c})$
    \State ${\epsilon}_{\text{uncond}} \leftarrow {\epsilon}_{\theta}^{\text{uncond}}(\boldsymbol{x}_k, k, \boldsymbol{c}_{\text{uncond}})$

    \State Extract $\mathbf{A}_{attn}$ and update clusters $\{\mathcal{C}_j\}_{j=1}^{K_c}$.
    \State Compute cluster-level scales $\lambda_{\mathcal{C}_j}$ by Eq.~\ref{eq:cluster_posterior}, Eq.~\ref{eq:cluster_scale}.

    \State Update $\lambda_i \leftarrow \lambda_{\mathcal{C}_i}$ for each node.

    \State $\boldsymbol{\lambda}_k \leftarrow (\lambda_1, \dots, \lambda_N)$
    \State $\tilde{\epsilon}_{\theta} \leftarrow {\epsilon}^{\text{uncond}}_{\theta} + \boldsymbol{\lambda}_k \odot ({\epsilon}^{\text{cond}}_{\theta} - {\epsilon}^{\text{uncond}}_{\theta})$
    \State Compute $\tilde{\mu}_{\theta}$ by $\tilde{\epsilon}_{\theta}$ and sample $\boldsymbol{x}_{k-1} \sim \mathcal{N}(\tilde{\mu}_{\theta}, \sigma_k^2\mathbf{I})$

    \State // Update posteriors for the next step
    \State Update $\log p_{\theta, k-1, i}(\boldsymbol{c}|\boldsymbol{x}_{k-1})$ using Eq.~\ref{eq:posterior_update_practical}, Eq.~\ref{eq:cluster_posterior}.
\EndFor
\State \textbf{return} $\boldsymbol{x}_0$
\end{algorithmic}
\end{algorithm}

\section{Experiments}

\subsection{Experimental Settings}
\subsubsection{Dateset.} We conduct experiments on the PEMS04, PEMS07, and PEMS08 datasets~\cite{chen2001freeway}. The datasets are split chronologically into training, validation, and test sets (60\%/20\%/20\%), and input samples are generated by segmenting these sets into overlapping sequences using a sliding window.

\subsubsection{Baselines.}
We evaluate the performance of FENCE against eight methods, covering machine learning baselines, discriminative models, and generative models. The discriminative models include: 1) \textbf{ASTGNN}~\cite{guo2021learning}, an attention-based graph neural network adapted for imputation via a reconstruction-based self-supervised learning objective~\cite{cao2018brits}; 2) \textbf{IGNNK}~\cite{wu2021inductive}, an inductive GNN for kriging; 3) \textbf{GCASTN}~\cite{peng2023generative}, a contrastive self-supervised learning framework for imputation; and 4) \textbf{ImputeFormer}~\cite{nie2024imputeformer}, which combines the Transformer with low-rank induction. The machine learning method is: 5) \textbf{LCR}~\cite{chen2024laplacian}, which leverages laplacian convolutional representations for time series imputation. The generative models include: 6) \textbf{mTAN}~\cite{shukla2021multi}, employing a VAE for irregularly sampled time series; 7) \textbf{CSDI}~\cite{tashiro2021csdi}, a conditional score-based diffusion model for imputation; and 8) \textbf{PriSTI}~\cite{liu2023pristi}, a conditional diffusion framework that integrates geographic context.

\subsubsection{Missing Patterns.}
We introduce two challenging missingness patterns: Spatially Random, Temporally Contiguous (SR-TC) and Spatially Clustered, Temporally Contiguous (SC-TC). 1) \textbf{SR-TC:} The total length of the series at each node is \(L\), and the time series is divided into $\frac{L}{T}$ non-overlapping temporal patches of length \(T\). For each of the \(N\) nodes, each temporal patch is independently masked with a probability of \(\alpha\), resulting in missing blocks that are contiguous in time but randomly distributed across nodes. 2) \textbf{SC-TC:} The \(N\) nodes are first partitioned into \(N_c\) distinct communities. A missing block is defined by a temporal patch of length \(T\) and a node community. Each of these $\frac{L}{T} \times N_c$ blocks is independently masked with a probability of \(\alpha\), causing entire communities of sensors to drop out simultaneously for continuous time periods. For our experiments, we set the missing rate \(\alpha = 80\%\) and \(T = 12\).

\subsection{Performance Comparison}
\setlength{\tabcolsep}{4.5pt} 

\begin{table*}[t]

\small % 表格内容保持小号字体（通常9pt）以适应版面
\centering

% --- 原来的 caption 位置已删除 ---

\renewcommand{\arraystretch}{1.25} 

\begin{tabular}{c|c|c|ccccc|ccc|c}
\toprule
Datasets & Miss Type & Metrics & ASTGNN & IGNNK & GCASTN & LCR & ImputeFormer & mTAN & CSDI & PriSTI & \textbf{FENCE} \\
\midrule
\multirow{6}{*}{PEMS04}
 & \multirow{3}{*}{SR-TC}
   & MAE      & 31.47 & 32.69 & 30.54 & 28.75 & \underline{27.30} & 31.20 & 27.63 & {27.51} & \textbf{26.57} \\
 & & RMSE     & 45.94 & 47.27 & 44.93 & 44.10 & {43.81} & 45.36 & \underline{43.37} & {43.46} & \textbf{42.45} \\
 & & MAPE     & 0.192 & 0.201 & 0.191 & 0.189 & \underline{0.178} & 0.209 & 0.187 & {0.184} & \textbf{0.172} \\
\cmidrule{2-12}
 & \multirow{3}{*}{SC-TC}
   & MAE      & 31.70 & 33.32 & 30.07 & 28.98 & {29.35} & 30.59 & \underline{28.26} & {28.47} & \textbf{27.31} \\
 & & RMSE     & 47.76 & 49.12 & 47.97 & 46.96 & {46.71} & 48.19 & {45.39} & \underline{45.31} & \textbf{44.28} \\
 & & MAPE     & 0.212 & 0.214 & 0.205 & 0.197 & {0.206} & 0.217 & \underline{0.189} & {0.190} & \textbf{0.180} \\
\midrule
\multirow{6}{*}{PEMS07}
 & \multirow{3}{*}{SR-TC}
   & MAE      & 46.16 & 52.64 & 50.02 & 47.24 & 45.07 & 45.60 & \underline{44.36} & 46.84 & \textbf{42.51} \\
 & & RMSE     & 65.60 & 71.16 & 66.31 & 65.88 & 65.86 & 67.07 & \underline{64.37} & 65.23 & \textbf{63.48} \\
 & & MAPE     & 0.206 & 0.234 & 0.310 & 0.207 & \underline{0.195} & 0.225 & 0.208 & 0.213 & \textbf{0.178} \\
\cmidrule{2-12}
 & \multirow{3}{*}{SC-TC}
   & MAE      & 47.63 & 55.12 & 45.95 & {44.97} & \underline{44.59} & 45.31 & 44.78 & 45.18 & \textbf{43.12} \\
 & & RMSE     & 73.85 & 79.54 & 74.29 & 74.11 & {73.75} & 75.06 & 74.33 & \underline{73.54} & \textbf{73.06} \\
 & & MAPE     & 0.265 & 0.332 & 0.260 & 0.248 & 0.232 & 0.247 & 0.228 & \underline{0.224} & \textbf{0.215} \\
\midrule
\multirow{6}{*}{PEMS08}
 & \multirow{3}{*}{SR-TC}
   & MAE      & {26.72} & 27.17 & 26.82 & 25.52 & {25.27} & 27.09 & 24.32 & \underline{24.06} & \textbf{22.77} \\
 & & RMSE     & 41.22 & 42.76 & 41.87 & 40.90 & \underline{40.64} & 44.92 & 41.09 & 42.01 & \textbf{40.26} \\
 & & MAPE     & 0.180 & 0.194 & 0.186 & 0.166 & \underline{0.160} & 0.187 & 0.167 & 0.164 & \textbf{0.147} \\
\cmidrule{2-12}
 & \multirow{3}{*}{SC-TC}
   & MAE      & 31.82 & 32.25 & 30.78 & 30.51 & {29.64} & 30.29 & \underline{28.23}  & {35.90} & \textbf{27.29} \\
 & & RMSE     & 50.76 & 51.35 & 49.39 & 50.03 & \underline{48.37} & 51.82 & 48.54 & 48.80 & \textbf{47.78} \\
 & & MAPE     & 0.220 & 0.231 & 0.213 & 0.218 & {0.216} & 0.229 & \underline{0.190} & 0.193 & \textbf{0.175} \\
\bottomrule
\end{tabular}

% --- 修改点：Caption 移到底部，并恢复字号 ---
\vspace{5pt} % 可选：增加一点表格和标题之间的间距
\normalsize  % 关键：将字体恢复为 10pt (Regular size)
\caption{Overall Imputation performance comparison. Bold and underlined fonts indicate the best and second-best results.}
\label{overall_comparison}

\end{table*}

The overall performance is presented in Tab.~\ref{overall_comparison}. The key observations are as follows: (1) FENCE achieves state-of-the-art performance across all three datasets, both for the SR-TC and SC-TC missing patterns. Notably, FENCE outperforms the second-best method by an average of \(6.26 \%\) in MAPE across all the datasets and missing patterns. (2) Among discriminative models, ImputeFormer demonstrates superior performance, benefiting from its integration of low-rank inductive bias combined with Transformers. (3) Score-based diffusion models, such as CSDI and PriSTI, perform competitively compared to machine learning and discriminative models. (4) FENCE significantly outperforms existing diffusion models, including CSDI and PriSTI, demonstrating the effectiveness of the dynamic feedback mechanism that adjusts the guidance scale during imputation. (5) Under the challenging SC-TC scenario, FENCE consistently outperforms all baselines across all metrics, demonstrating its effectiveness in handling highly sparse missing patterns.

\subsection{Ablation Study}
We conduct an ablation study to evaluate the effectiveness of the spatial-temporal feedback guidance mechanism and the cluster-aware guidance strategy. Specifically, we compare FENCE with three variants: (1) \textbf{wo-U}, which removes the modeling of unconditional scores and only models the conditional scores. (2) \textbf{wo-F}, which removes the spatial-temporal feedback guidance from the denoising process; (3) \textbf{wo-C}, which removes the cluster-aware guidance strategy and instead applies a uniform global guidance scale to all nodes at each denoising step.

\begin{figure}[h]
    \centering
    \includegraphics[width= \linewidth]{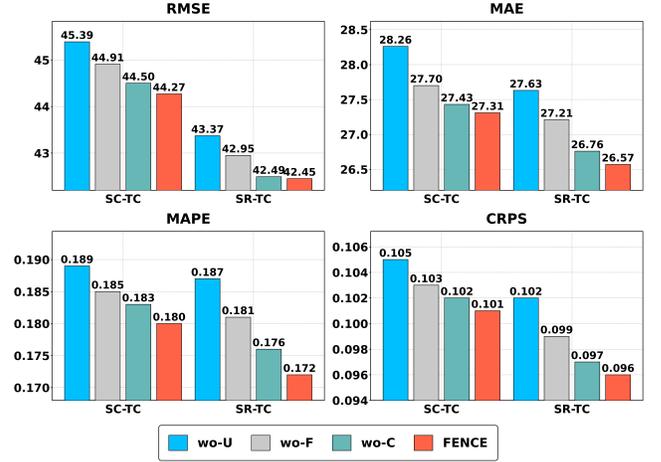}
      \caption{Ablation study.}
  \label{fig:ablation}  
\end{figure}

The results are shown in Fig.~\ref{fig:ablation}. First, the \textbf{wo-U} variant, which does not model the unconditional prior distribution, results in suboptimal performance. This indicates that the two-stage training procedure, which first models the prior distribution, facilitates a more accurate modeling of the conditional distribution. Second, we compare FENCE with the \textbf{wo-F} variant. The results show that incorporating spatial-temporal feedback guidance yields substantial performance gains across all metrics and missing patterns, showing the importance of dynamically adjusting the guidance scale based on the alignment between generated values and conditional observations. Finally, FENCE outperforms the \textbf{wo-C} variant, showing the effectiveness of the cluster-aware guidance strategy. This result demonstrates that computing guidance scales based on clustered information leads to more accurate imputations, compared to applying a uniform global scale to all nodes at each denoising step.

\subsection{Hyperparameter Analysis}
\begin{figure}[ht]
    \centering
    \includegraphics[width= \linewidth]{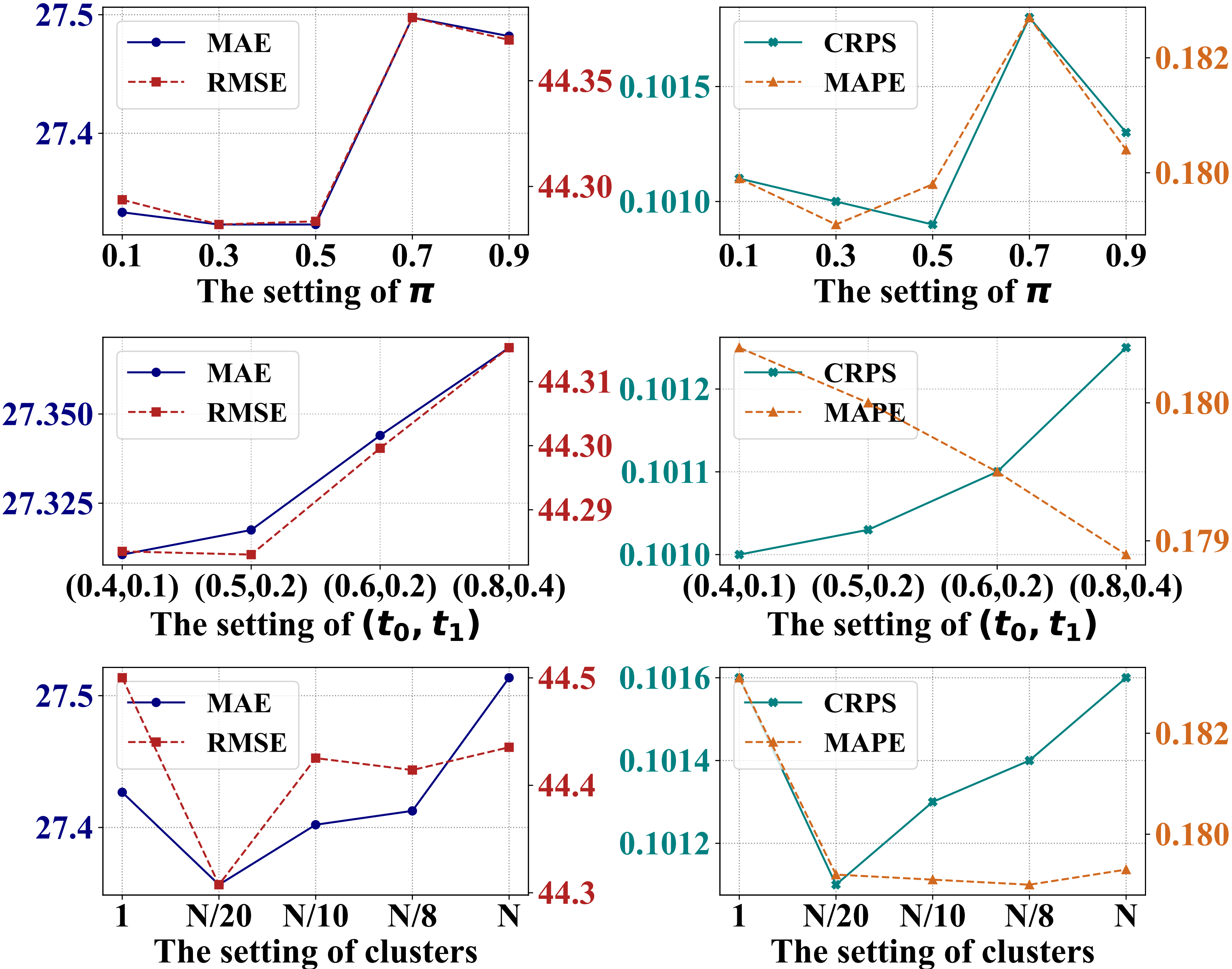}
      \caption{Effect of hyperparameters.}
  \label{fig:hyperparam}  
\end{figure}

We evaluate the key hyperparameters of FENCE on the PEMS04 dataset, as shown in Fig.~\ref{fig:hyperparam}. These include the prior confidence $\pi$, the guidance timing parameters $(t_0, t_1)$, and the number of clusters. The parameters $(t_0, t_1)$ control the guidance offset $\delta$ and temperature $\tau$, with their relationships illustrated in the appendix.
As shown in Fig.~\ref{fig:hyperparam}, FENCE's performance is relatively stable across different settings of $(t_0, t_1)$. Regarding the prior confidence in the conditional model, the best results are obtained when $\pi = 0.5$. A higher value of $\pi$ indicates high confidence in the conditional model, which necessitates a very low posterior likelihood for applying guidance. In contrast, a lower value, such as $\pi=0.5$, provides a broader operational range for FENCE, lowering the threshold for guidance activation. Next, we evaluate different settings of the ratio of node number to cluster number: \(1, N/20, N/10, N/8, N\). The best performance is achieved at \(N/20\), while setting the ratio to \(1\) or \(N\) results in degraded performance. This indicates that using either a global uniform guidance scale or a node-specific guidance scale is suboptimal compared to employing a cluster-level guidance scale.

\subsection{Case Study}

\begin{figure}[H]
    \centering
    \includegraphics[width= \linewidth]{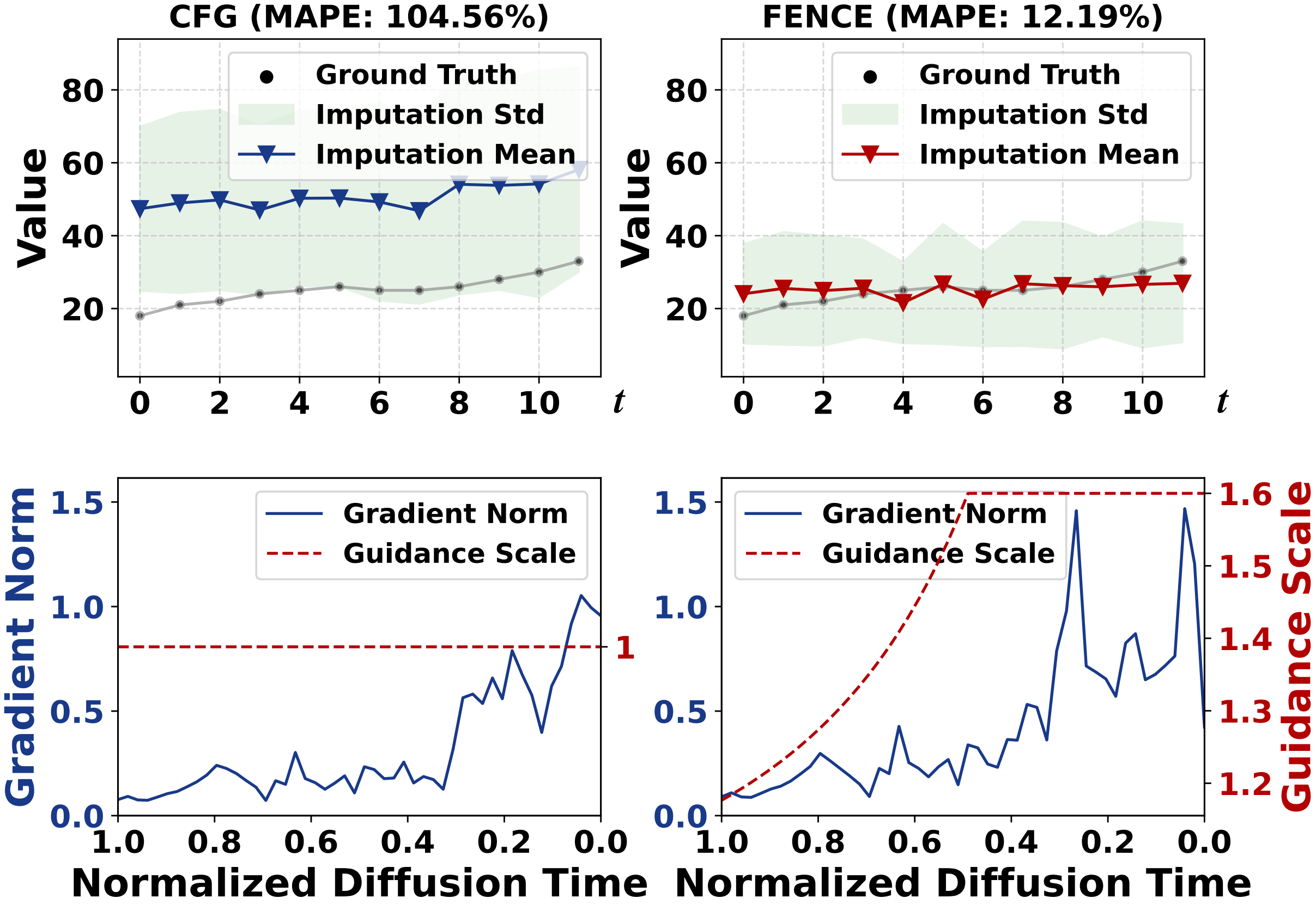}
      \caption{Case Study}
  \label{fig:case_study}  
\end{figure}

We compare FENCE with CFG, where the guidance scale is fixed at 1. In Fig.~\ref{fig:case_study}, for a node with no observations across 12 time slices, CFG's fixed scale provides insufficient correction strength, causing the imputation to revert to the learned average and deviate from the ground truth. In contrast, FENCE's dynamic guidance mechanism actively adjusts the guidance scale based on the posterior likelihood. When the imputation diverges from the observations, the guidance scale increases to strengthen the correction. This adaptive process results in an imputation that more accurately reflects the true conditional data distribution.

\section{Conclusion}
This paper proposes FENCE, a spatial-temporal feedback diffusion guidance method that tackles the limitations of existing imputation methods based on diffusion models, which rely on a fixed guidance scale. FENCE dynamically adjusts the guidance scale based on posterior likelihood approximations, ensuring the generated values consistently align with observed data throughout the denoising process. Furthermore, the cluster-aware guidance mechanism leverages spatial-temporal correlations to tailor the guidance for different nodes, improving imputation accuracy.

\section{Acknowledgments}
This work was supported by the National Natural Science Foundation of China (No. 62372031) and the Beijing Natural Science Foundation (Grant No. 4242029).

\bibliography{aaai2026}

\newpage
\section{Appendix}
This appendix provides supplementary details for our work. We begin by presenting mathematical derivations for the guidance scale and the posterior likelihood update rule. Following the derivations, we describe the practical implementation of hyper-parameters. Finally, we introduce our experimental settings in detail, including hyper-parameter configurations, the computing environment, evaluation metrics, and the datasets.

\subsection{Derivation of the Guidance Scale} 

This section provides a detailed derivation for the guidance scale $\lambda(\boldsymbol{x}_k, k)$, which is applied at each denosing step \(k\) in the formulation of the guidance score:

\begin{equation}
\begin{split}
    \nabla_{\boldsymbol{x}_k} \log \tilde{p}_{k}(\boldsymbol{x}_{k}|\boldsymbol{c}) 
    &= s_{\theta}(\boldsymbol{x}_{k}) \\
     \quad &+ \lambda(\boldsymbol{x}_{k}, k) \left( s_{\theta}(\boldsymbol{x}_k, \boldsymbol{c}) - s_{\theta}(\boldsymbol{x}_{k}) \right)
\end{split}
\label{eq:fbg_form}
\end{equation}
The derivation begins with the additive error assumption introduced in~\cite{koulischer2025feedback}. The assumption states that the learned distribution is a linear combination of the true conditional distribution $p_k(\boldsymbol{x}_k|\boldsymbol{c})$ and the true unconditional distribution $p_k(\boldsymbol{x}_k)$:
\begin{equation}
    p_{\theta,k}(\boldsymbol{x}_{k}|\boldsymbol{c}) = (1 - \pi)p_{k}(\boldsymbol{x}_{k}) + \pi p_{k}(\boldsymbol{x}_{k}|\boldsymbol{c})
    \label{eq:additive_error_assumption}
\end{equation}
where $\pi \in [0, 1]$ is a hyper-parameter representing the prior confidence in how well the learned model approximates the true conditional distribution. Rearranging Eq. \ref{eq:additive_error_assumption} yields:
\begin{equation}
    \pi p_{k}(\boldsymbol{x}_{k}|\boldsymbol{c}) = p_{\theta,k}(\boldsymbol{x}_{k}|\boldsymbol{c}) - (1 - \pi)p_{k}(\boldsymbol{x}_{k})
    \label{eq: cond_uncond}
\end{equation}
This equation contains the true unconditional distribution $p_k(\boldsymbol{x}_k)$, which is unknown. To make the formulation practical, we assume that the model’s learned unconditional distribution approximates the true one:
\begin{equation}
    p_{\theta,k}(\boldsymbol{x}_k) \approx p_k(\boldsymbol{x}_k)
\end{equation}
Substituting this approximation into Eq.~\ref{eq: cond_uncond} yields:
\begin{equation}
    \pi p_{k}(\boldsymbol{x}_{k}|\boldsymbol{c}) \approx p_{\theta,k}(\boldsymbol{x}_{k}|\boldsymbol{c}) - (1 - \pi)p_{\theta,k}(\boldsymbol{x}_{k})
\end{equation}
Since $\pi$ is a positive constant, this relationship can be interpreted as a proportionality. We define the approximated conditional distribution $\tilde{p}_{k}(\boldsymbol{x}_{k}|\boldsymbol{c})$ to be proportional to the true conditional distribution $p_{k}(\boldsymbol{x}_{k}|\boldsymbol{c})$, resulting in:
\begin{equation}
    \tilde{p}_{k}(\boldsymbol{x}_{k}|\boldsymbol{c}) \propto p_{\theta,k}(\boldsymbol{x}_{k}|\boldsymbol{c}) - (1 - \pi)p_{\theta,k}(\boldsymbol{x}_{k})
    \label{eq:app_target_dist}
\end{equation}
This expression defines an approximated conditional distribution in terms of the distributions the model has learned. 

Next, we compute the score function $\nabla_{\boldsymbol{x}_k} \log \tilde{p}_{k}(\boldsymbol{x}_{k}|\boldsymbol{c})$ based on this proportionality. To do so, we first rewrite the proportionality relationship in Eq.~\ref{eq:app_target_dist} as an equality by introducing a normalization constant $Z$:

\begin{equation}
    \tilde{p}_{k}(\boldsymbol{x}_{k}|\boldsymbol{c}) = \frac{1}{Z} \left( p_{\theta,k}(\boldsymbol{x}_{k}|\boldsymbol{c}) - (1 - \pi)p_{\theta,k}(\boldsymbol{x}_{k}) \right)
    \label{eq: Z_p}
\end{equation}
Here, $Z$ is the normalization constant and is defined as:
\begin{equation}
    Z = \int \left( p_{\theta,k}(\boldsymbol{x}_{k}'|\boldsymbol{c}) - (1 - \pi)p_{\theta,k}(\boldsymbol{x}_{k}') \right) d\boldsymbol{x}_{k}'
    \nonumber
\end{equation}
Next, we take the logarithm of both sides of the equality in Eq.~\ref{eq: Z_p}:
\begin{align}
    \log \tilde{p}_{k}(\boldsymbol{x}_{k}|\boldsymbol{c}) &= \log \left( \frac{1}{Z} \left( p_{\theta,k}(\boldsymbol{x}_{k}|\boldsymbol{c}) - (1 - \pi)p_{\theta,k}(\boldsymbol{x}_{k}) \right) \right) \nonumber \\
    % Second line with line break
    &= -\log(Z) \nonumber \\
     \qquad &+ \log \left( p_{\theta,k}(\boldsymbol{x}_{k}|\boldsymbol{c}) - (1 - \pi)p_{\theta,k}(\boldsymbol{x}_{k}) \right)
    \label{eq: log_x_k_derive}
\end{align}

We then compute the gradient of the expression defined in Eq.~\ref{eq: log_x_k_derive} with respect to $\boldsymbol{x}_k$:
\begin{align}
    \nabla_{\boldsymbol{x}_k} \log \tilde{p}_{k}(\boldsymbol{x}_{k}|\boldsymbol{c}) & = \nabla_{\boldsymbol{x}_k} \left( -\log(Z) + \log \left( p_{\theta,k}(\boldsymbol{x}_{k}|\boldsymbol{c}) \right. \right. \nonumber \\
     \phantom{=}& \left. \left. - (1 - \pi)p_{\theta,k}(\boldsymbol{x}_{k}) \right) \right)
\end{align}
Applying the sum rule for derivatives:

\begin{align}
    \nabla_{\boldsymbol{x}_k} \log \tilde{p}_{k}(\boldsymbol{x}_{k}|\boldsymbol{c})
    & = \nabla_{\boldsymbol{x}_k}(-\log Z) \nonumber \\
     \phantom{=} &+ \nabla_{\boldsymbol{x}_k} \log \Bigl( p_{\theta,k}(\boldsymbol{x}_{k}|\boldsymbol{c}) \nonumber \\
     \phantom{= + \nabla_{\boldsymbol{x}_k} \log \bigl(}& - (1 - \pi)p_{\theta,k}(\boldsymbol{x}_{k}) \Bigr)
\end{align}
Since \(Z\) is a constant, its logarithm \(\log Z\) is also constant, and thus its gradient is zero. As a result, the gradient simplifies to:
\begin{multline}
    \nabla_{\boldsymbol{x}_k} \log \tilde{p}_{k}(\boldsymbol{x}_{k}|\boldsymbol{c}) = \\
    \nabla_{\boldsymbol{x}_k} \log \left( p_{\theta,k}(\boldsymbol{x}_{k}|\boldsymbol{c}) - (1 - \pi)p_{\theta,k}(\boldsymbol{x}_{k}) \right)
\end{multline}
Applying the identity $\nabla \log f(\boldsymbol{x}) = \frac{\nabla f(\boldsymbol{x})}{f(\boldsymbol{x})}$ yields the formulation:
\begin{equation}
    \nabla_{\boldsymbol{x}_k} \log \tilde{p}_{k}(\boldsymbol{x}_{k}|\boldsymbol{c}) = \frac{\nabla_{\boldsymbol{x}_k} \left( p_{\theta,k}(\boldsymbol{x}_{k}|\boldsymbol{c}) - (1 - \pi)p_{\theta,k}(\boldsymbol{x}_{k}) \right)}{p_{\theta,k}(\boldsymbol{x}_{k}|\boldsymbol{c}) - (1 - \pi)p_{\theta,k}(\boldsymbol{x}_{k})}
    \label{eq:app_score_raw}
\end{equation}
To introduce the posterior term $p_{\theta,k}(\boldsymbol{c}|\boldsymbol{x}_k)$, we apply Bayes' theorem, $p(\boldsymbol{x}|\boldsymbol{c}) = \frac{p(\boldsymbol{c}|\boldsymbol{x})p(\boldsymbol{x})}{p(\boldsymbol{c})}$, to the terms in Eq.~\ref{eq:app_score_raw}. We begin by rewriting the denominator:
\begin{align}
    \text{Denominator} &= p_{\theta,k}(\boldsymbol{x}_k|\boldsymbol{c}) - (1 - \pi)p_{\theta,k}(\boldsymbol{x}_k) \nonumber \\
    &= \frac{p_{\theta,k}(\boldsymbol{c}|\boldsymbol{x}_k)p_{\theta,k}(\boldsymbol{x}_k)}{p_{\theta,k}(\boldsymbol{c})} - (1 - \pi)p_{\theta,k}(\boldsymbol{x}_k) \nonumber \\
    &= p_{\theta,k}(\boldsymbol{x}_k) \left( \frac{p_{\theta,k}(\boldsymbol{c}|\boldsymbol{x}_k)}{p_{\theta,k}(\boldsymbol{c})} - (1 - \pi) \right)
    \label{eq:app_denominator}
\end{align}
Next, we rewrite the term inside the gradient in the numerator:
\begin{equation}
\begin{split}
    p_{\theta,k}(\boldsymbol{x}_k|\boldsymbol{c})  &- (1 - \pi)p_{\theta,k}(\boldsymbol{x}_k) = \\
    & p_{\theta,k}(\boldsymbol{x}_k) \left(
    \frac{p_{\theta,k}(\boldsymbol{c}|\boldsymbol{x}_k)}{p_{\theta,k}(\boldsymbol{c})}
    - (1 - \pi)
    \right)
\end{split}
\end{equation}
We compute the gradient of the numerator using the gradient product rule:
\begin{align}
    \text{Numerator} 
    &= \nabla_{\boldsymbol{x}_k} \left[ p_{\theta,k}(\boldsymbol{x}_k) \left( \frac{p_{\theta,k}(\boldsymbol{c}|\boldsymbol{x}_k)}{p_{\theta,k}(\boldsymbol{c})} - (1 - \pi) \right) \right] \nonumber \\
    &= \left( \frac{p_{\theta,k}(\boldsymbol{c}|\boldsymbol{x}_k)}{p_{\theta,k}(\boldsymbol{c})} - (1 - \pi) \right) \nabla_{\boldsymbol{x}_k} p_{\theta,k}(\boldsymbol{x}_k) \nonumber \\
     \qquad &+ p_{\theta,k}(\boldsymbol{x}_k) \nabla_{\boldsymbol{x}_k} \left( \frac{p_{\theta,k}(\boldsymbol{c}|\boldsymbol{x}_k)}{p_{\theta,k}(\boldsymbol{c})} \right)
\end{align}
Using the score identity, $\nabla_{\boldsymbol{x}_k} p(\boldsymbol{x}_k) = p(\boldsymbol{x}_k) \nabla_{\boldsymbol{x}_k} \log p(\boldsymbol{x}_k) = p(\boldsymbol{x}_k) s_{\theta}(\boldsymbol{x}_k)$, we substitute the gradient of the unconditional distribution into the numerator expression:
\begin{multline}
    \text{Numerator} = \left( \frac{p_{\theta,k}(\boldsymbol{c}|\boldsymbol{x}_k)}{p_{\theta,k}(\boldsymbol{c})} - (1 - \pi) \right) p_{\theta,k}(\boldsymbol{x}_k) s_{\theta}(\boldsymbol{x}_k) \\
    + p_{\theta,k}(\boldsymbol{x}_k) \nabla_{\boldsymbol{x}_k} \left( \frac{p_{\theta,k}(\boldsymbol{c}|\boldsymbol{x}_k)}{p_{\theta,k}(\boldsymbol{c})} \right)
    \label{eq:app_numerator}
\end{multline}
Substituting the formulations for the numerator (Eq.~\ref{eq:app_numerator}) and denominator (Eq.~\ref{eq:app_denominator}) into Eq.~\ref{eq:app_score_raw} yields:

\begin{align}
    &\nabla_{\boldsymbol{x}_k} \log \tilde{p}_{k}(\boldsymbol{x}_{k}|\boldsymbol{c}) 
    = \frac{1}{p_{\theta,k}(\boldsymbol{x}_k) \left( \frac{p_{\theta,k}(\boldsymbol{c}|\boldsymbol{x}_k)}{p_{\theta,k}(\boldsymbol{c})} - (1 - \pi) \right)} \nonumber \\
    & \qquad \times \Biggl[ \left( \frac{p_{\theta,k}(\boldsymbol{c}|\boldsymbol{x}_k)}{p_{\theta,k}(\boldsymbol{c})} - (1 - \pi) \right) p_{\theta,k}(\boldsymbol{x}_k) s_{\theta}(\boldsymbol{x}_k) \nonumber \\
     \qquad & \qquad + p_{\theta,k}(\boldsymbol{x}_k) \nabla_{\boldsymbol{x}_k} \left( \frac{p_{\theta,k}(\boldsymbol{c}|\boldsymbol{x}_k)}{p_{\theta,k}(\boldsymbol{c})} \right) \Biggr] \nonumber \\
    &= s_{\theta}(\boldsymbol{x}_k) + \frac{\nabla_{\boldsymbol{x}_k} \left( p_{\theta,k}(\boldsymbol{c}|\boldsymbol{x}_k) / p_{\theta,k}(\boldsymbol{c}) \right)}{p_{\theta,k}(\boldsymbol{c}|\boldsymbol{x}_k) / p_{\theta,k}(\boldsymbol{c}) - (1 - \pi)}
    \label{eq:app_intermediate_score}
\end{align}
Using the score identity again, and noting that 
\(\nabla_{\boldsymbol{x}_k} \log p(\boldsymbol{c}|\boldsymbol{x}_k) = s_{\theta}(\boldsymbol{x}_k, \boldsymbol{c}) - s_{\theta}(\boldsymbol{x}_k)\), we obtain:
\begin{align}
    \nabla_{\boldsymbol{x}_k} \left( \frac{p_{\theta,k}(\boldsymbol{c}|\boldsymbol{x}_k)}{p_{\theta,k}(\boldsymbol{c})} \right) &= \frac{1}{p_{\theta,k}(\boldsymbol{c})} \nabla_{\boldsymbol{x}_k} p_{\theta,k}(\boldsymbol{c}|\boldsymbol{x}_k) \nonumber \\
    &= \frac{p_{\theta,k}(\boldsymbol{c}|\boldsymbol{x}_k)}{p_{\theta,k}(\boldsymbol{c})} \nabla_{\boldsymbol{x}_k} \log p_{\theta,k}(\boldsymbol{c}|\boldsymbol{x}_k) \nonumber \\
    &= \frac{p_{\theta,k}(\boldsymbol{c}|\boldsymbol{x}_k)}{p_{\theta,k}(\boldsymbol{c})} \left( s_{\theta}(\boldsymbol{x}_k, \boldsymbol{c}) - s_{\theta}(\boldsymbol{x}_k) \right)
\end{align}
Substituting this result into Eq.~\ref{eq:app_intermediate_score} yields:
\begin{align}
    \nabla_{\boldsymbol{x}_k} \log \tilde{p}_{k}(\boldsymbol{x}_{k}|\boldsymbol{c}) 
    &= s_{\theta}(\boldsymbol{x}_k) \nonumber \\
    & \quad + \left( \frac{p_{\theta,k}(\boldsymbol{c}|\boldsymbol{x}_k) / p_{\theta,k}(\boldsymbol{c})}{p_{\theta,k}(\boldsymbol{c}|\boldsymbol{x}_k) / p_{\theta,k}(\boldsymbol{c}) - (1 - \pi)} \right) \nonumber \\
    & \quad \times \left( s_{\theta}(\boldsymbol{x}_k, \boldsymbol{c}) - s_{\theta}(\boldsymbol{x}_k) \right)
    \label{eq:score_derive}
\end{align}
By comparing Eq.~\ref{eq:score_derive} with the guidance score formulation in Eq.~\ref{eq:fbg_form}, we can identify the guidance scale $\lambda(\boldsymbol{x}_k, k)$:
\begin{equation}
    \lambda(\boldsymbol{x}_k, k) = \frac{p_{\theta,k}(\boldsymbol{c}|\boldsymbol{x}_k) / p_{\theta,k}(\boldsymbol{c})}{p_{\theta,k}(\boldsymbol{c}|\boldsymbol{x}_k) / p_{\theta,k}(\boldsymbol{c}) - (1 - \pi)}
\end{equation}
In practice, assuming the prior $p_{\theta,k}(\boldsymbol{c})$ is constant yields:
\begin{equation}
    \lambda(\boldsymbol{x}_k, k) = \frac{p_{\theta,k}(\boldsymbol{c}|\boldsymbol{x}_k)}{p_{\theta,k}(\boldsymbol{c}|\boldsymbol{x}_k) - (1-\pi)}
    \label{eq:lambda_practical}
\end{equation}
This completes the derivation of guidance scale.

\subsection{Derivation of the Posterior Likelihood Estimation} 

In this section, we derive the posterior likelihood estimation employed in the feedback guidance mechanism:
\begin{multline} \label{eq:posterior_update_practical}
    \log p_{\theta}(\boldsymbol{c}|\boldsymbol{x}_{k-1}) = \log p_{\theta}(\boldsymbol{c}|\boldsymbol{x}_k) \\
    - \frac{\tau}{2\sigma_k^2} \Bigl( \|\boldsymbol{x}_{k-1} - {\mu}_{\theta}(\boldsymbol{x}_k|\boldsymbol{c})\|^2 - \|\boldsymbol{x}_{k-1} - {\mu}_{\theta}(\boldsymbol{x}_k)\|^2 \Bigr) - \delta.
\end{multline}
We aim to derive an update rule for estimating \(\log p(\boldsymbol{c}|\boldsymbol{x}_{k-1})\) based on its value at the previous step, \(\log p(\boldsymbol{c}|\boldsymbol{x}_k)\). The derivation begins with the application of the chain rule of probability to the posterior \(p(\boldsymbol{c}|\boldsymbol{x}_{k-1:K})\). By exploiting the Markov property of the reverse diffusion process, where the distribution of \(\boldsymbol{x}_{k-1}\) depends only on \(\boldsymbol{x}_k\), we use the conditional independence \(
p(\boldsymbol{x}_{k-1}|\boldsymbol{x}_{k:K}, \boldsymbol{c}) = p(\boldsymbol{x}_{k-1}|\boldsymbol{x}_k, \boldsymbol{c}) \) to establish a relationship between the posterior at step \(k - 1\) and the posterior at step \(k\):
\begin{equation}
    p(\boldsymbol{c}|\boldsymbol{x}_{k-1}) = p(\boldsymbol{c}|\boldsymbol{x}_k) \cdot \frac{p(\boldsymbol{x}_{k-1}|\boldsymbol{x}_k, \boldsymbol{c})}{p(\boldsymbol{x}_{k-1}|\boldsymbol{x}_k)}
    \label{eq:bayesian_post}
\end{equation}
Taking the logarithm of Eq.~\ref{eq:bayesian_post} yields the log-likelihood update rule:
\begin{equation}
\begin{split}
    \log p(\boldsymbol{c}|\boldsymbol{x}_{k-1}) 
    &= \log p(\boldsymbol{c}|\boldsymbol{x}_k) \\
    & \quad + \log p(\boldsymbol{x}_{k-1}|\boldsymbol{x}_k, \boldsymbol{c}) \\
    & \quad - \log p(\boldsymbol{x}_{k-1}|\boldsymbol{x}_k)
\end{split}
\label{eq:log_posterior_update}
\end{equation}
We model the reverse transition processes as Gaussian distributions. The unconditional reverse transition is given by 
\(p_\theta(\boldsymbol{x}_{k-1}|\boldsymbol{x}_k) \sim \mathcal{N}(\boldsymbol{x}_{k-1}; \boldsymbol{\mu}_\theta(\boldsymbol{x}_k), \sigma_k^2 \mathbf{I})\), 
while the conditional reverse transition is modeled as 
\(p_\theta(\boldsymbol{x}_{k-1}|\boldsymbol{x}_k, \boldsymbol{c}) \sim \mathcal{N}(\boldsymbol{x}_{k-1}; \boldsymbol{\mu}_\theta(\boldsymbol{x}_k|\boldsymbol{c}), \sigma_k^2 \mathbf{I})\). 
Here, \(\boldsymbol{\mu}_\theta(\boldsymbol{x}_k)\) and \(\boldsymbol{\mu}_\theta(\boldsymbol{x}_k|\boldsymbol{c})\) denote the means predicted by the unconditional and conditional models, respectively, while the variance \(\sigma_k^2\) is assumed to be fixed across both processes.

The log-probability density function of a Gaussian distribution $\mathcal{N}(\boldsymbol{x}; \boldsymbol{\mu}, \sigma^2 \mathbf{I})$ is given by $-\frac{1}{2\sigma^2}\|\boldsymbol{x} - \boldsymbol{\mu}\|^2$ plus a constant term that does not depend on the mean $\boldsymbol{\mu}$. When computing the difference between the conditional and unconditional log-likelihoods, this constant term cancels out. 
The resulting difference is:
\begin{equation}
\begin{split}
    & \log p(\boldsymbol{x}_{k-1}|\boldsymbol{x}_k, \boldsymbol{c}) - \log p(\boldsymbol{x}_{k-1}|\boldsymbol{x}_k) \\
    & \qquad = \left(-\frac{1}{2\sigma_k^2}\|\boldsymbol{x}_{k-1} - \boldsymbol{\mu}_\theta(\boldsymbol{x}_k|\boldsymbol{c})\|^2\right) \\
     \qquad & \qquad - \left(-\frac{1}{2\sigma_k^2}\|\boldsymbol{x}_{k-1} - \boldsymbol{\mu}_\theta(\boldsymbol{x}_k)\|^2\right) 
\end{split}
\label{eq:log_likelihood_diff}
\end{equation}
This formulation allows the posterior likelihood to be updated by comparing the outputs of the conditional and unconditional models at each step. Accordingly, Eq.~\ref{eq:log_posterior_update} becomes:
\begin{equation}
\begin{split}
    \log p_{\theta}(\boldsymbol{c}|\boldsymbol{x}_{k-1}) 
    &= \log p_{\theta}(\boldsymbol{c}|\boldsymbol{x}_k) \\
     \quad & + \frac{1}{2\sigma_k^2} \Biggl( \|\boldsymbol{x}_{k-1} - \boldsymbol{\mu}_\theta(\boldsymbol{x}_k)\|^2 \\
     \qquad &- \|\boldsymbol{x}_{k-1} - \boldsymbol{\mu}_\theta(\boldsymbol{x}_k|\boldsymbol{c})\|^2 \Biggr)
\end{split}
\end{equation}
Additionally, we introduce two hyper-parameters: a temperature $\tau$ to scale the update strength and an offset $\delta$ to ensure guidance activates properly in early diffusion stages. This results in the following parameterized update equation for the posterior:
\begin{multline} \label{eq:posterior_update_practical}
    \log p_{\theta}(\boldsymbol{c}|\boldsymbol{x}_{k-1}) = \log p_{\theta}(\boldsymbol{c}|\boldsymbol{x}_k) \\
    - \frac{\tau}{2\sigma_k^2} \Bigl( \|\boldsymbol{x}_{k-1} - {\mu}_{\theta}(\boldsymbol{x}_k|\boldsymbol{c})\|^2 - \|\boldsymbol{x}_{k-1} - {\mu}_{\theta}(\boldsymbol{x}_k)\|^2 \Bigr) - \delta.
\end{multline}
This completes the derivation of the posterior likelihood update rule.

\subsection{Practical Implementation of Hyper-parameters} 
To simplify control of the guidance during the denoising process, we replace direct tuning of the hyper-parameters \(\tau\) and \(\delta\) with two denoising time-based parameters: \(t_0\) and \(t_1\). These correspond to specific points in the denoising process, where \(t = 0\) represents the clean data and \(t = 1\) corresponds to pure noise. By specifying these time points, we enable adjustment of the guidance strength throughout the denoising process.

The guidance strength can be modulated over time by three key parameters: \(t_0\), \(t_1\), and \(\pi\). The activation time \(t_0\) determines the point in the denoising process at which the guidance scale reaches the predefined reference value \(\lambda_{ref}\). The peak time \(t_1\) represents the time at which the guidance strength reaches its maximum, after which its influence begins to decrease. As the noised data becomes cleaner during the denoising process, the influence of guidance gradually diminishes. The peak time \(t_1\) is used to compute the overall scaling factor of the guidance strength, referred to as the temperature \(\tau\). Finally, the prior confidence factor \(\pi\) is a time-invariant parameter that specifies the relative weighting between the model’s conditional and unconditional outputs. These parameters provide a flexible way to modulate guidance strength throughout the denoising process.

The relationship between these parameters is established through a two-step calculation. First, the offset \(\delta\) is computed based on the specified time \(t_0\) and the prior confidence factor \(\pi\). This step ensures that guidance becomes effective during the early stages of the denoising process. The parameter \(\lambda_{ref}\) is a predefined reference value of the guidance scale:
\begin{align}
\delta &= \frac{1}{(1 - t_0)\,K}
  \log\!\biggl(\frac{(1 - \pi)\,\lambda_{ref}}{\lambda_{ref} - 1}\biggr)
\end{align}
Second, the temperature parameter \(\tau\) is calculated based on the specified time \(t_1\), using the noise variance \(\sigma_{t_1}^2\) at that time and the previously computed offset \(\delta\). This step determines the overall scaling of the guidance strength across the denoising process:
\begin{align}
\tau &= \left\lvert \frac{2\,\sigma_{t_1}^2}{\alpha_{\mathrm{scale}}} \, \delta \right\rvert
\end{align}
Here, \(K\) denotes the total number of denoising steps, and \(\alpha_{\mathrm{scale}}\) is an empirically chosen scaling factor, typically set to 10.

In summary, this reparameterization simplifies the tuning of hyperparameters \((\tau, \delta)\) by formulating it as the task of defining a guidance schedule. This is achieved by specifying the time \((t_0)\) at which guidance scale reaches a predefined value, and the time (\(t_1)\) at which the guidance strength reaches its maximum, after which it begins to decrease.

\subsection{Experimental Settings}

\subsubsection{Hyper-parameter Configurations}
Baselines are implemented following the parameters suggested in their original papers. For the hyper-parameters of FENCE, the batch size is 128. For unconditional generation, we train for 150 epochs with a learning rate of 2e-3 and apply early stopping with a patience of 20 epochs. For conditional generation, we train for 80 epochs with a learning rate of 1e-3 and apply early stopping with a patience of 10 epochs. The diffusion model uses a minimum noise level $\beta_1$ and a maximum noise level $\beta_K$, where K is set to 50. We adopt the quadratic schedule for the intermediate noise levels, formalized as
\[
\beta_k = \left(\frac{K - k}{K - 1}\sqrt{\beta_1} + \frac{k - 1}{K - 1}\sqrt{\beta_K}\right)^2
\]
The diffusion time embedding and temporal encoding are implemented by sine and cosine embeddings. We summarize the hyper-parameters of FENCE in Table~\ref{tab:optimal-parameter}. All experiments are run five times.

\begin{table}[t]
  \centering
  \caption{The hyper-parameters of FENCE.}
  \label{tab:optimal-parameter}
  \begin{tabular}{@{}ll@{}}
    \toprule
    \textbf{Description}               & \textbf{Values}                                        \\
    \midrule
    Batch size                        & 128                                                     \\
    Length of time slices $L$                   & 12                  \\
    Layers of noise estimation        & 4                                                      \\
    Channel size $d$                  & 64                                                     \\
    Number of attention heads         & 8                                                      \\
    Diffusion embedding dim           & 64                                                      \\
    Time embedding dim                & 128      \\
    Feature embedding dim             & 16           \\
    Minimum noise level $\beta_1$     & 0.0001                                                 \\
    Maximum noise level $\beta_K$     & 0.5                                                    \\
    Diffusion steps $K$               & 50                \\
    Prior confidence $\pi$             & 0.5             \\
    Reference guidance scale $\lambda_{ref}$      & 1.6           \\
    Activation time \(t_0\) & 0.8 \\
    Peak time \(t_1\)    & 0.5   \\
    \bottomrule
  \end{tabular}
\end{table}

\subsubsection{Environmental Settings}
All methods are implemented using Python and PyTorch. 
% For the hyper-parameters of the proposed method, we consider six key parameters in different modules; their range and optimal values are listed in Table~\ref{tab:optimal-parameter}, where $L_T$ represents the number of layers for the transformer used in path prediction, $E_T$ denotes the dimension of the hidden state in the transformer for path prediction, and $E_U$ denotes the dimension of the hidden state in MoEUQ.
During training, two Adam optimizers with weight‐decay coefficients of 1e-6 and 1e-5 are employed for the unconditional and conditional diffusion models, each governed by a MultiStepLR scheduler that decays the learning rate by a factor of 0.1 at 75 \% and 90 \% of the total training epochs.
We run all experiments on Ubuntu 20.04 servers equipped with Intel(R) Xeon(R) W-2155 CPUs and NVIDIA GPUs (RTX A4000 and RTX 3090).

\subsubsection{Evaluation metrics} To quantitatively evaluate the imputation accuracy of all models, we employ three commonly adopted metrics to assess mean error: Mean Absolute Error (MAE), Root Mean Squared Error (RMSE), and Mean Absolute Percentage Error (MAPE).
For generative models such as FENCE and its diffusion-based counterparts, the final deterministic imputation is obtained by averaging 10 generated samples from the learned distribution before calculating the metrics.

In our ablation studies and hyper-parameter analysis, we additionally employ the Continuous Ranked Probability Score (CRPS) to assess the quality of the entire predictive distribution.  For a missing value \(x\) with estimated distribution \(D\), CRPS is defined as
\begin{align}
\mathrm{CRPS}\bigl(D^{-1}, x\bigr)
&= \int_{0}^{1} 2\,\Lambda_{\alpha}\bigl(D^{-1}(\alpha), x\bigr)\,\mathrm{d}\alpha,
\label{eq:crps-continuous}\\
\Lambda_{\alpha}\bigl(D^{-1}(\alpha), x\bigr)
&= \bigl(\alpha - \mathbf{1}_{\,x < D^{-1}(\alpha)}\bigr)\,\bigl(x - D^{-1}(\alpha)\bigr),
\label{eq:quantile-loss}
\end{align}
where \(\alpha\in[0,1]\) is the quantile level, \(D^{-1}(\alpha)\) is the \(\alpha\)-quantile of \(D\), and \(\mathbf{1}\) is the indicator function.  In practice, we approximate the integral by sampling 100 draws from \(D\) and computing
\begin{equation}
\mathrm{CRPS}\bigl(D^{-1}, x\bigr)
\approx \frac{1}{19}
\sum_{i=1}^{19}
2\,\Lambda_{\,i\times 0.05}\bigl(D^{-1}(i\times 0.05), x\bigr),
\label{eq:crps-discrete}
\end{equation}
and then average across all missing entries \(\bar X\) to get
\begin{equation}
\mathrm{CRPS}\bigl(D, \bar X\bigr)
= \frac{1}{|\bar X|}
\sum_{x\in\bar X}
\mathrm{CRPS}\bigl(D^{-1}, x\bigr).
\label{eq:crps-mean}
\end{equation}

\subsubsection{Datasets}

We conduct experiments on three real-world traffic datasets: PEMS04, PEMS07, and PEMS08. These datasets are part of the Caltrans Performance Measurement System~\cite{chen2001freeway} and provide data aggregated at 5-minute intervals.
For the imputation task, we exclusively utilizes the traffic flow feature from each dataset and normalize missing values using the global mean and standard deviation computed from the fully observed training data.
Detailed statistics of these datasets are in Tab.~\ref{data-description}.
\setlength{\tabcolsep}{7pt}
\begin{table}[h]
\centering
\caption{Dataset Description.}
\label{data-description}
\begin{tabular}{cccc}
\toprule
 \multicolumn{1}{c}{\bf Properties}  &\multicolumn{1}{c}{\bf PEMS04} &\multicolumn{1}{c}{\bf PEMS07}  &\multicolumn{1}{c}{\bf PEMS08}  \\
\midrule
Time range     & 2 months &  4 months & 2 months  \\
Time interval  & 5 min    &  5 min & 5 min             \\
\# of nodes    & 307      &  883 & 170        \\
Mean of flow   & 207      &  65 & 229       \\
Std of flow    & 156      &  41 & 145     \\
\bottomrule
\end{tabular}
\end{table}

\end{document}